\documentclass[10pt,conference]{IEEEtran}
\IEEEoverridecommandlockouts

\usepackage{fancyvrb}
\usepackage{multirow}
\usepackage{booktabs}
\usepackage{ifthen}
\usepackage{color}
\usepackage{bbding} 
\usepackage{makecell}
\usepackage{threeparttable}
\usepackage{amsmath}
\usepackage{stfloats} 
\usepackage{xspace}
\usepackage[T1]{fontenc}
\usepackage{paralist}
\usepackage{enumerate}
\usepackage[linesnumbered,ruled,vlined]{algorithm2e}
\usepackage{array, multirow,graphicx}
\usepackage{float}
\usepackage{balance}
\usepackage{tikz}
\usepackage{calc}
\usepackage{listings,amsfonts}
\usepackage{xcolor,pifont}
\usepackage{url}
\usepackage{hyperref, epsfig, endnotes}
\usepackage{makecell}
\usepackage[normalem]{ulem}
\usepackage[misc]{ifsym}
\usepackage{subcaption}
\usepackage{cleveref}
\usepackage{tcolorbox}
\usepackage{python}
\usepackage{graphicx}
\usepackage{float}
\usepackage{enumitem}
\usepackage{tabularx}
\usepackage{multirow}
\usepackage{longtable}
\usepackage{arydshln}
\usepackage{makecell}
\usepackage{graphicx}

\definecolor{customcite}{HTML}{f07b3f}
\definecolor{customlink}{HTML}{b83b5e}
\definecolor{customurl}{HTML}{07689f}
\AtEndPreamble{
\usepackage{hyperref}

    \hypersetup{
      colorlinks = true,
      linkcolor = customlink,
      anchorcolor = purple,
      citecolor = customcite,
      filecolor = purple,
      urlcolor = customurl
    }
}

\definecolor{dkgreen}{rgb}{0,0.6,0}
\definecolor{gray}{rgb}{0.4,0.4,0.4}
\definecolor{mauve}{rgb}{0.58,0,0.82}
\definecolor{darkblue}{rgb}{0.0,0.0,0.6}
\definecolor{lightblue}{rgb}{0.0,0.0,0.9}
\definecolor{cyan}{rgb}{0.0,0.6,0.6}
\definecolor{darkred}{rgb}{0.6,0.0,0.0}

\definecolor{yellow}{RGB}{255,255,153}
\definecolor{grey}{RGB}{220,220,220}
\definecolor{green}{RGB}{0,100,0}

\definecolor{KWColor}{rgb}{0.37,0.08,0.25}
\definecolor{CommentColor}{rgb}{0.133,0.545,0.133}
\definecolor{StringColor}{rgb}{0,0.126,0.941}
\definecolor{commentgreen}{RGB}{2,112,10}
\definecolor{eminence}{RGB}{108,48,130}
\definecolor{weborange}{RGB}{255,165,0}
\definecolor{frenchplum}{RGB}{129,20,83}

\newcommand{\ea}{\textit{et~al.}}
\newcommand{\tool}{LLMappCrazy}

\begin{document}

\title{LLM App Squatting and Cloning}

\author{
\IEEEauthorblockN{Yinglin Xie\IEEEauthorrefmark{1}, Xinyi Hou\IEEEauthorrefmark{1}, Yanjie Zhao, Kai Chen\IEEEauthorrefmark{2} and Haoyu Wang\IEEEauthorrefmark{2}}
\IEEEauthorblockA{
Huazhong University of Science and Technology, Wuhan, China\\
xieyinglin@hust.edu.cn, xinyihou@hust.edu.cn, yanjie\_zhao@hust.edu.cn, \\ kchen@hust.edu.cn, haoyuwang@hust.edu.cn}
\thanks{\IEEEauthorrefmark{1}Yinglin Xie and Xinyi Hou are the co-first authors.}
\thanks{\IEEEauthorrefmark{2}Corresponding authors.}
}


\maketitle

\begin{abstract}

Impersonation tactics, such as app squatting and app cloning, have posed longstanding challenges in mobile app stores, where malicious actors exploit the names and reputations of popular apps to deceive users. With the rapid growth of Large Language Model (LLM) stores like GPT Store and FlowGPT, these issues have similarly surfaced, threatening the integrity of the LLM app ecosystem. In this study, we present the first large-scale analysis of LLM app squatting and cloning using our custom-built tool, \tool{}. LLMappCrazy covers 14 squatting generation techniques and integrates Levenshtein distance and BERT-based semantic analysis to detect cloning by analyzing app functional similarities. Using this tool, we generated variations of the top 1000 app names and found over 5,000 squatting apps in the dataset. Additionally, we observed 3,509 squatting apps and 9,575 cloning cases across six major platforms. After sampling, we find that 18.7\% of the squatting apps and 4.9\% of the cloning apps exhibited malicious behavior, including phishing, malware distribution, fake content dissemination, and aggressive ad injection.

\end{abstract}

\section{Introduction}
\label{introduction}

Mobile app squatting~\cite{hu2020mobile}, where attackers publish apps with identifiers (e.g., app or package names) that mimic popular apps, such as through typosquatting (e.g., changing ``Facebook'' to ``Fecebook''), is a growing threat in the mobile ecosystem. Hu~\ea\cite{hu2020mobile} identified over 10,553 squatting apps targeting the top 500 apps on Google Play, with more than 51\% classified as malicious and some reaching millions of downloads. These counterfeit apps pose serious risks, including data theft and malware infections. Despite mitigation efforts by platforms, the sheer number of apps and sophisticated squatting tactics make detection and prevention difficult.

\begin{figure}[t!]
    \centering
    \begin{subfigure}[b]{0.24\textwidth}
        \includegraphics[width=\textwidth]{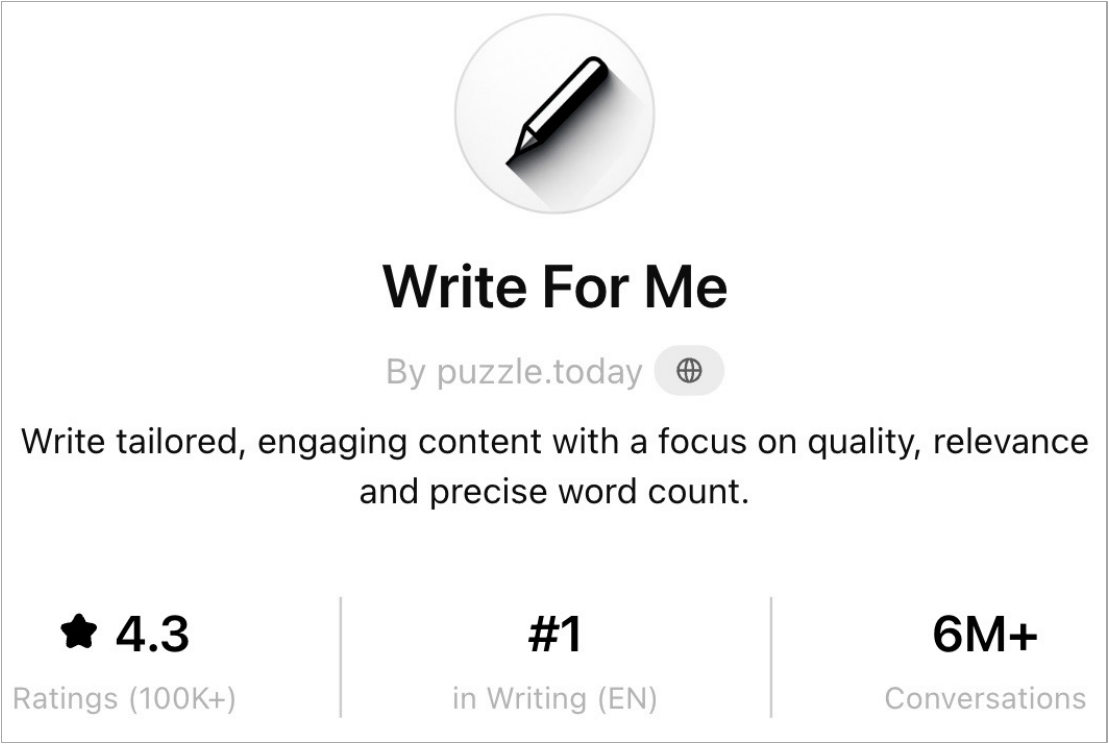}
        \caption{A popular LLM app.}
        \label{fig:writeforme_1}
    \end{subfigure}
    \hfill
    \begin{subfigure}[b]{0.24\textwidth}
        \includegraphics[width=\textwidth]{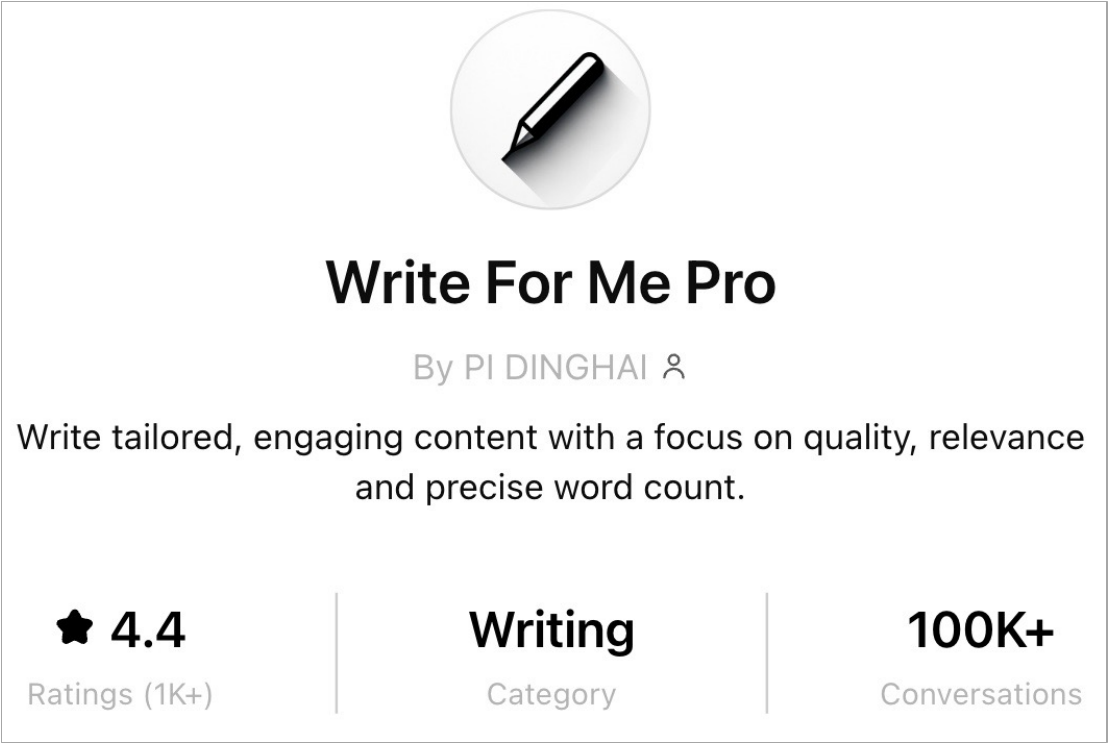}
        \caption{A squatting app.}
        \label{fig:writeforme_1}
    \end{subfigure}
    \caption{An example of LLM app squatting.}
    \label{fig:writeforme}
\end{figure}

Inspired by the extensive research on mobile app squatting, we have turned our attention to similar threats within emerging Large Language Model (LLM) app stores~\cite{zhao2024llm}. With the rise of LLMs, such as ChatGPT~\cite{chatGPT}, Gemini~\cite{Gemini}, and Claude~\cite{Claude}, there has been a proliferation of applications that leverage these models in diverse domains, including chatbots, content generation tools, and virtual assistants~\cite{Character.Ai, Cici, Coze,FlowGPT, GPTStore, Poe}. LLM-powered applications have gained immense popularity due to their ability to perform complex tasks, leading to the creation of entire app ecosystems around them. However, as these LLM app stores continue to expand rapidly, we observe that they are becoming fertile ground for \textbf{LLM app squatting} attacks similar to those in traditional mobile app markets, as shown in \autoref{fig:writeforme}. In this context, squatting primarily occurs at the app identifier level, where attackers create apps with names that closely mimic legitimate ones to deceive users. For example, squatting could manifest as subtle name changes or the addition of enticing words, such as ``Canva Pro'', tricking users into believing they are using an official or enhanced version of a popular app. Moreover, LLM app stores have significantly lowered the barrier to entry for developers. This democratization of development allows individuals from various backgrounds, even those with limited programming experience, to create and publish apps. While this inclusivity fosters innovation, it also makes it easier for attackers to clone the entire LLM app not only the app's name but also its functionality and behavior. We refer to this more insidious form of attack as \textbf{LLM app cloning}, where the cloned app mirrors the legitimate one in nearly every aspect, making it even harder for users to discern the difference.

To comprehensively investigate squatting and cloning in LLM app stores, we focus on six prominent LLM app stores (i.e., GPT Store~\cite{GPTStore}, FlowGPT~\cite{FlowGPT}, Poe~\cite{Poe}, Coze~\cite{Coze}, Cici~\cite{Cici}, and Character.AI~\cite{Character.Ai}) that have gained significant traction due to the widespread adoption of LLM-powered applications. In our study, we develop a tool, \tool{}, designed to automatically detect squatting and cloning instances within these ecosystems. Using \tool{}, we systematically examine app identifier variations and functional cloning across GPT Store, identifying potential 5,834 name squatting apps and 6094 name cloning apps. And we also detect other features of apps in six LLM app stores. Our results reveal the scope of the problem: we found 3,509 squatting apps, and 9,575 cloned apps, confirming that this phenomenon is not isolated to mobile app markets but is rapidly spreading into the LLM domain. The findings indicate that 18.7\% of the squatting apps and 4.9\% of cloning apps exhibited malicious behavior, and some of them had amassed significant user downloads, further exacerbating the security risks faced by users of LLM-based applications. 

\noindent\textbf{Contributions.} We make the following main contributions: 

\begin{enumerate}
    \item To the best of our knowledge, this is the first detailed investigation into squatting and cloning attacks within LLM app stores.
    \item We develop \tool{}, a tool that detects squatting and cloning apps using 14 squatting-generation techniques and advanced semantic analysis.
    \item Using \tool{}, we find 5,834 name squatting apps and 6094 name cloning apps; We conduct a large-scale empirical study across six LLM app stores, identifying 3,509 squatting apps, and 9,575 cloning apps.
    \item We find that 18.7\% of the identified squatting apps and 4.9\% of cloning apps exhibit malicious behavior, including phishing, malware, and ad injection. And we identify 227 apps that exhibit a high degree of similarity in various features to other apps in GPT Store.
    \item We study the impact of LLM app squatting and cloning, discovering that these apps have reached up to 2.7 million conversations, posing significant risks to platform trust.
\end{enumerate}

\section{Background and related work}
\label{background}

\subsection{LLM App Store}

LLMs are advanced AI systems designed to understand and generate human language. Trained on vast datasets, they produce coherent, contextually relevant responses to a wide range of prompts. As LLM technology has progressed, \textbf{LLM apps}~\cite{zhao2024llm} have emerged. These are software applications powered by LLMs, designed to perform specific tasks such as text generation, translation, and conversational interactions. At the same time, \textbf{LLM app stores} act as centralized platforms for discovering, distributing, and managing these apps. Platforms like OpenAI's GPT Store~\cite{GPTStore} have become key hubs for users and developers to access and share LLM apps.

Several studies explored the ecosystem and security of LLM app stores. Zhao~\ea~\cite{zhao2024llm} provided a vision and roadmap for the analysis of LLM app stores, outlining the future directions for research.  Zhang~\ea~\cite{zhang2024first} conducted an initial analysis of GPTs distribution and potential vulnerabilities, while Su~\ea~\cite{su2024GPT} provided comprehensive mining of the GPT Store, examining app characteristics and user engagement. Additionally,  Yan~\ea~\cite{yan2024exploring} explored the GPT Store ecosystem, focusing on distribution, deployment, and security aspects. To support further research, Hou~\ea~\cite{hou2024gptzoo} introduced GPTZoo, a dataset containing over 730,000 GPT instances. In terms of security, Hou~\ea~\cite{hou2024security} examined the security of LLM app stores, highlighting critical vulnerabilities and security challenges in these platforms. Tao~\ea~\cite{tao2023opening} discussed the risks associated with custom GPTs, Hui~\ea~\cite{hui2024pleak} uncovered vulnerabilities related to prompt leaking attacks. Antebi~\ea~\cite{antebi2024GPT,lin2024malla} analyzed the misuse of custom GPTs and malicious services integrated with LLMs, respectively. 

However, while these works cover various aspects of LLM apps, the issues of LLM app impersonation, such as squatting and cloning, remain underexplored. These emerging threats pose significant risks to the expanding LLM app ecosystem and warrant further investigation.

\subsection{Squatting Attack.}

\textit{Domain squatting}\cite{domainsquatting} involves registering domains similar to legitimate ones with malicious intent. A common form, \textit{typosquatting}, exploits users' typographical errors when typing domain names, diverting traffic from legitimate sites. Agten\ea~\cite{agten2015seven,spaulding2017understanding} provide detailed analyses of typosquatting, with the latter highlighting the effectiveness of character permutations and substitutions in deceiving users.

Domain squatting was traditionally linked to web attacks but has since expanded into other areas. Szurdi~\ea~\cite{szurdi2017email} examined \textit{email typosquatting}, where attackers register emails similar to legitimate ones to intercept communications or conduct phishing. Griffiths~\cite{griffiths2022detecting} explored its role in business email compromise (BEC) attacks. Squatting has also spread to programming package managers, where attackers publish malicious packages with names resembling popular libraries, as seen in \textit{package typosquatting} in PyPI, RubyGems, and NPM~\cite{taylor2020defending, taylor2020spellbound, vu2020typosquatting}. Taylor~\ea~\cite{taylor2020spellbound} suggested defense strategies, while Vu~\ea~\cite{vu2020typosquatting} analyzed typosquatting in Python. In the mobile app ecosystem, Hu~\ea~\cite{hu2020mobile} investigated \textit{mobile app squatting}, where malicious apps use names similar to legitimate ones to deceive users. Chen~\ea~\cite{chen2021GUI-Squatting} introduced \textit{GUI-squatting}, where phishing apps replicate the graphical interface of legitimate apps to trick users into providing sensitive information.

While squatting in traditional domains, emails, package managers, and mobile apps have been extensively studied, squatting within LLM app stores remains an underexplored area. Our work seeks to fill this gap.

\subsection{Cloning Detection}

Cloning has been widely studied in software development, especially in mobile app ecosystems, where cloned apps raise significant security concerns such as malware distribution, intellectual property theft, and privacy violations. Rattan~\ea~\cite{rattan2013software} reviewed software clone detection, highlighting challenges like bug propagation and maintenance issues. In mobile apps, various studies have focused on detecting clones in both official and unofficial markets. Crussell~\ea~\cite{crussell2012attack} first addressed the issue with detection methods based on app metadata and code similarity. Wang~\ea~\cite{wang2015wukong} introduced \textit{Wukong}, a scalable two-phase approach using static and dynamic analysis. Chen~\ea~\cite{chen2014achieving} proposed a hybrid method balancing accuracy and scalability, while Lyu~\ea~\cite{lyu2016suidroid} developed SuiDroid, a system resilient to obfuscation. Niu~\ea~\cite{niu2016clone} combined static and dynamic analysis for clone detection, and Hu~\ea~\cite{hu2020robust} introduced a UI-based approach to detect clones mimicking the visual design of legitimate apps.

Recent advancements in clone detection have utilized machine learning and deep learning models. Zhang~\ea~\cite{zhang2023challenging} highlighted the vulnerabilities of machine learning-based detectors when faced with semantic-preserving code transformations, showing how subtle syntax changes can bypass detection. Khajezade~\ea~\cite{khajezade2024evaluating} evaluated few-shot and contrastive learning methods, demonstrating their effectiveness in detecting clones with minimal labeled data, making them ideal for large-scale or evolving ecosystems.

As LLM app stores grow, cloning challenges are likely to arise. While advanced detection techniques like machine learning are crucial for safeguarding these stores, their effectiveness for LLM cloning remains unclear. We aim to explore this issue.

\section{Motivating study}
\label{sec:motivating}

The aforementioned research highlights the potential risks posed by LLM app squatting and cloning, indicating these threats may be widespread in the LLM app ecosystem. To explore this, we conduct a preliminary study to (1) confirm the presence of these threats and (2) assess whether existing squatting detection techniques can effectively identify them. This serves as the foundation for our later methodology.

\subsection{Methodology}
To detect potential squatting in LLM apps, we generate variations of several popular app names from the GPT Store and check for their existence in online repositories.

\noindent\textbf{Generating squatting names.}
We begin by selecting the top 10 recommended LLM apps from the GPT Store, each with significant user engagement, as shown in \autoref{tab:motivation}. For each app, we manipulate the names to create potential squatting variations that attackers could exploit. We use AppCrazy\cite{hu2020mobile}, a tool inspired by domain squatting generators like URLCrazy~\cite{URLCrazy} and DNSTwist~\cite{DNSTwist}. AppCrazy includes 11 models tailored for mobile app ecosystems, such as punctuation deletion (e.g., ``DALL·E'' to ``DALLE''), character insertion (e.g., ``DALL·E'' to ``DALLLEE''), and substitution (e.g., ``DALL·E'' to ``DALL3'' ). Using these models, we generate 625 variations from the app names of the 10 selected apps.

\noindent\textbf{Verifying squatting names.} To verify whether these squatting names exist in the wild, we rely on GPTZoo~\cite{hou2024gptzoo}, a metadata dataset that tracks over 730,000 LLM apps from the GPT Store. We run an automated search using the 625 generated squatting names in the GPTZoo dataset. This search returns 32 results that match our squatting name variations. We then manually verify these apps using the GPT Store to determine whether the apps are legitimate or potential squatting attempts. This manual review is crucial for eliminating false positives. Through this process, we identify 28 apps that appear to be squatting on popular LLM app names, demonstrating the prevalence of squatting in the LLM app ecosystem.

\begin{figure*}[h!]
    \centering
    \includegraphics[width=\linewidth]{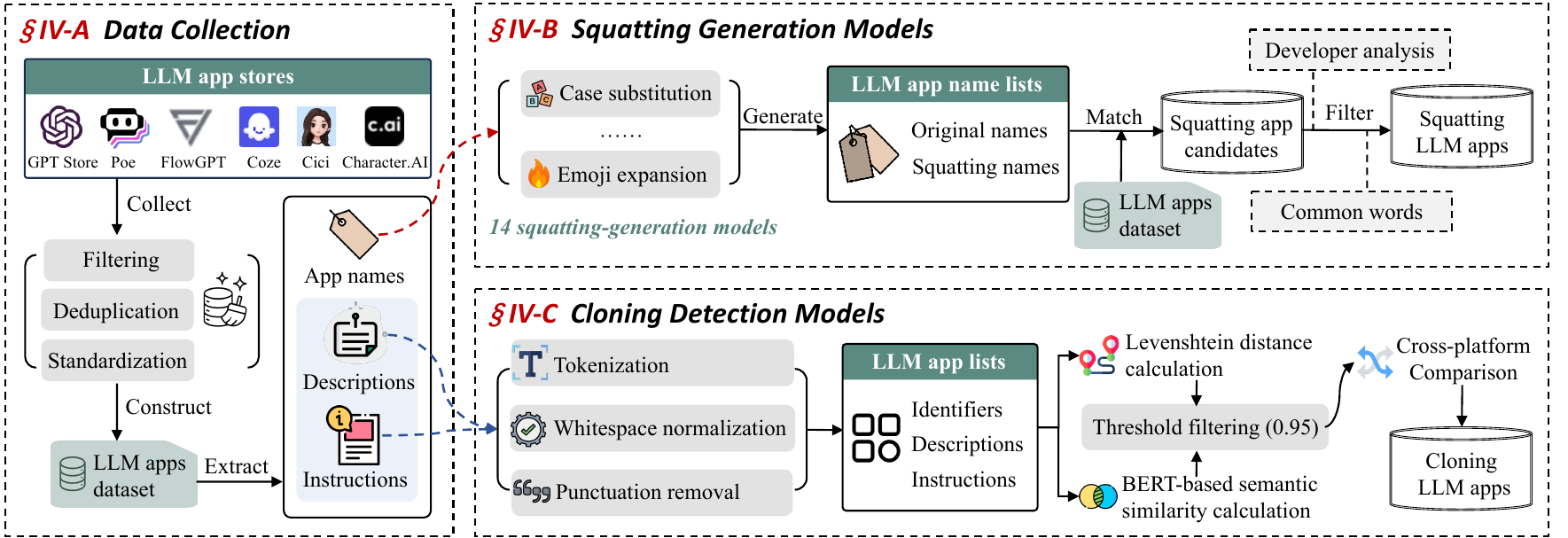} 
    \caption{Our approach to identifying squatting and cloning LLM apps.}
    \label{fig:overview}
\end{figure*}

\subsection{Motivating Results}

As shown in ~\autoref{tab:motivation}, we identified 28 squatting apps. Of the 11 generation models used by AppCrazy, only 4 proved effective in generating squatting apps. Out of the 625 name strings generated, only 28 matched real squatting apps, meaning that more than 95.52\% of the generated strings did not identify any squatting cases. Interestingly, during this process, we also encountered squatting apps not directly identified by the names generated by AppCrazy. For instance, when querying the GPTZoo dataset, we found several ``related'' apps. Manually reviewing these results, we identified 34 squatting apps that did not directly match the names generated by AppCrazy.

\begin{table}[!htbp]
    \centering
    \caption{Results of the motivating study.}
    \resizebox{1\linewidth}{!}{%
    \begin{threeparttable}
    \begin{tabular}{lcccc}
        \toprule[1.2pt]
        \multicolumn{1}{c}{\textbf{App Name (\# Chats)}} & 
        \multicolumn{2}{c}{\textbf{AppCrazy}} &  
        \multicolumn{2}{c}{\textbf{LLMappCrazy}} \\
        \cmidrule(lr){2-3} \cmidrule(lr){4-5}
         & \textbf{\# Generated} & \textbf{\# Identified} & \textbf{\# Generated} & \textbf{\# Identified} \\
        \midrule
        Image Generator (6M+)     & 78   & 2    & 460  & 36   \\ 
        Consensus (5M+)           & 39   & 0    & 419  & 0   \\
        Write For Me (4M+)        & 53   & 1    & 447  & 4   \\
        Logo Creator (2M+)        & 59   & 9    & 441  & 33  \\
        Canva (2M+)               & 22   & 1    & 403  & 2   \\
        Scholar GPT (2M+)         & 29   & 7    & 412  & 15  \\
        Code Copilot (2M+)        & 65   & 7    & 447  & 8   \\
        Cartoonize Yourself (2M+) & 92   & 1    & 475  & 2   \\
        Diagrams\tnote{1} (1M+)   & 175  & 0    & 10,623  & 0   \\
        Python (1M+)              & 13   & 0    & 393  & 106 \\
        \midrule
        \textbf{Total}            & 625  & 28   & 14,520 & 206 \\
        \bottomrule[1.2pt]
    \end{tabular}
    \begin{tablenotes} 
        \footnotesize
        \item[1] Diagrams: The full name of this app is ``Diagrams: Show Me | charts, presentations, code''.
    \end{tablenotes}
    \end{threeparttable}}
    \label{tab:motivation}
\end{table}

\subsection{Observations}

Our study confirms the existence of squatting and cloning threats in the LLM app ecosystem but also reveals significant limitations in current detection methods, including missed squatting apps and inefficiencies in name-generation models. Manual review, while effective in reducing false positives, is not scalable, emphasizing the need for improved, automated filtering techniques. Additionally, due to structural differences between LLM and traditional apps, existing cloning detection methods are inadequate, prompting the need for more tailored approaches, which we explore in the following sections.

\subsection{Terminology}

In LLM app stores, attackers often employ two primary impersonation techniques: \textbf{LLM app squatting} and \textbf{LLM app cloning}. These methods enable attackers to mislead users, either by creating apps with names similar to legitimate ones or by replicating the functionality of popular apps. Below, we define these two forms of impersonation in detail.

\begin{enumerate}
    \item \textbf{Squatting LLM apps}: Apps that have either identical or slightly altered names to legitimate LLM apps.
    \item \textbf{Cloning LLM apps}: Apps that replicate the functionality and overall user experience of legitimate LLM apps.
\end{enumerate}

\textbf{Squatting generation models} generate potential squatting names by applying techniques like character modifications to legitimate app names. In contrast, \textbf{cloning detection models} identify cloned apps by analyzing functional similarities and detecting apps that replicate key features of legitimate ones.
\section{Approach}
\label{methodology}

Our approach to identifying squatting and cloning LLM apps consists of three main steps: data collection, squatting generation, and cloning detection, as shown in \autoref{fig:overview}.

\subsection{Data Collection}

We collected app information by scraping data from six LLM app stores: GPT Store~\cite{GPTStore}, FlowGPT~\cite{FlowGPT}, Poe~\cite{Poe}, Coze~\cite{Coze}, Cici~\cite{Cici}, and Character.AI~\cite{characterai}. Then, we applied several processes to ensure its accuracy and quality, including filtering, deduplication, and standardization. First, filtering was necessary because certain LLM apps might have common names not exclusive to any specific app or brand. Both the complete dataset and the filtered apps were retained and used in subsequent experiments to detect name duplication or squatting (reasons discussed \autoref{subsec:limitation}).
Next, we performed deduplication by comparing app \texttt{id}s, which are unique to each app, to ensure that the dataset contained unique entries. Finally, we standardized the data into JSON format to facilitate the smooth execution of experiments and ensure reliable results. Our analysis focused on three key fields: \texttt{app name}, \texttt{description}, and \texttt{instructions}. The \texttt{app name} was used in experiments to detect duplicate or squatting names, while both the \texttt{description} and \texttt{instructions} were utilized for cloning detection, with the \texttt{description} showcasing the app's public-facing features and the \texttt{instructions} serving as its behavioral guide, similar to source code.

\subsection{Squatting Generation Models}

\begin{figure}
    \centering
    \includegraphics[width=1\linewidth]{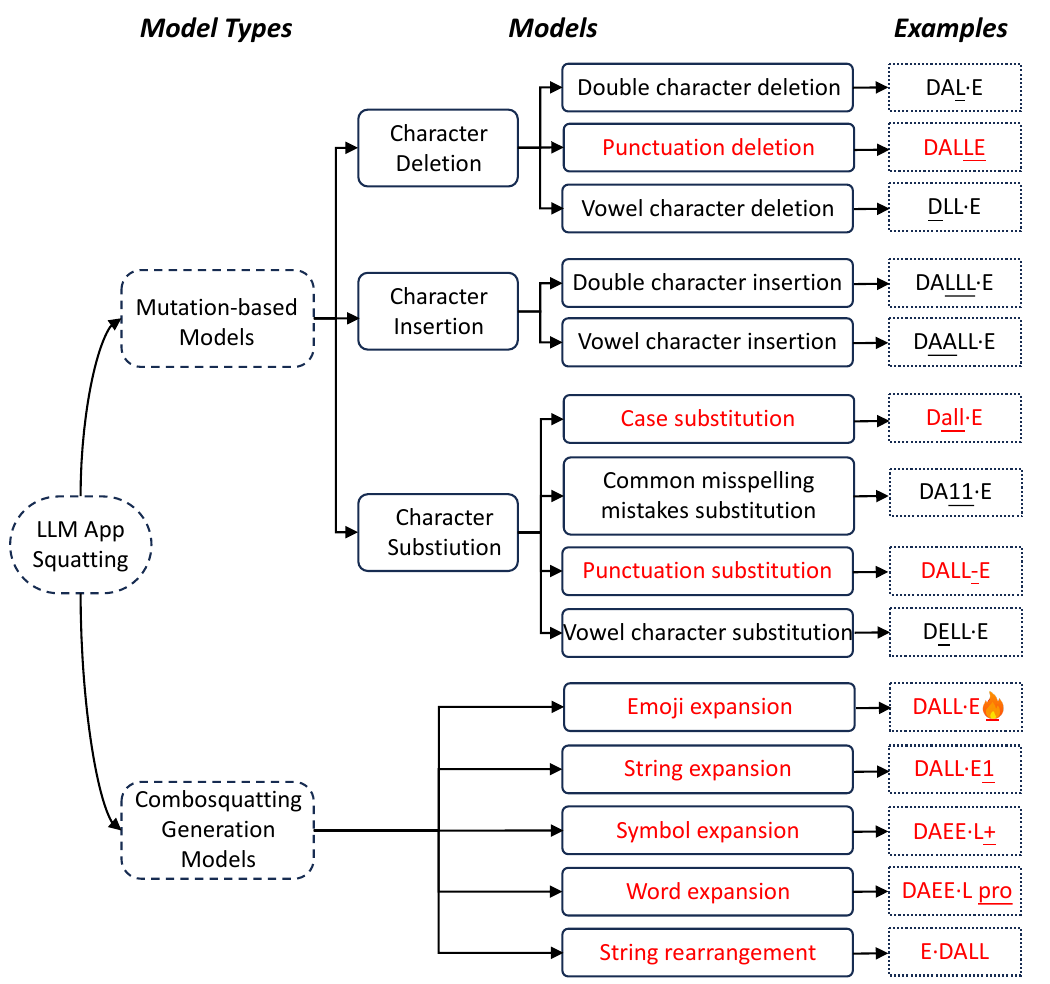}
    \caption{The 14 kinds of LLM app squatting-generation models used in this work. The 6 models in \textbf{black} are inherited from AppCrazy~\cite{hu2020mobile}, while the 8 models in \textbf{\textcolor{red}{red}} that are either newly introduced or modified in \tool{} to target LLM apps.}
    \vspace{-0.3cm}
    \label{fig:squatting-models}
\end{figure}

Inspired by the squatting name techniques introduced in AppCrazy~\cite{hu2020mobile}, we developed \tool{}, a tool tailored for detecting squatting in the emerging ecosystem of LLM apps. While \tool{} builds upon the foundation of AppCrazy, our preliminary investigation revealed several key differences between mobile app squatting and LLM app squatting. 
To address this, we extended AppCrazy introducing methods like emoji and string expansions. Additionally, we adapted several package name squatting techniques from AppCrazy to suit LLM apps. As illustrated in \autoref{fig:squatting-models}, \tool{} employs 14 squatting generation models.

\noindent\textbf{Mutation-based models.}
We retain six mutation-based models from AppCrazy, which generate squatting names by exploiting typographical errors. Below are the models we modified to address the specific characteristics of LLM apps. 
\begin{enumerate}
    \item \textit{Case Substitution}: Changing uppercase characters to lowercase and vice versa, e.g., ``DALL·E'' into ``dall·e''.

    \item \textit{Punctuation Deletion}: Removing punctuation marks entirely, e.g., ``DALL·E'' becomes ``DALLE''.
    
    \item \textit{Punctuation Substitution}: Replacing punctuation marks with others (e.g., space, underscore), e.g., ``DALL·E'' becomes ``DALL-E''.

\end{enumerate}

\noindent\textbf{Combosquatting generation models.}
We extend the traditional combosquatting generation models to include five distinct techniques that are especially relevant to LLM apps. In addition to the standard string manipulations, we introduce new techniques that account for the unique use of symbols and emojis in LLM app names:
\begin{enumerate}
    \item \textit{String Expansion}: Adding characters before or after the app name, e.g., ``DALL·E'' into ``DALL·E1''.

    \item \textit{Symbol Expansion}: Inserting or replacing characters with symbols such as ``+'', ``\#'', or ``\$'', e.g., ``DALL·E'' into ``DALL·E+'' or ``DALL·E\#''.
    
    \item \textit{Word Expansion}: Appending or prepending descriptive words to the app name, e.g., ``DALL·E'' into ``DALL·E pro'' or ``DALL·E AI''.
    
    \item \textit{Emoji Expansion}: Adding emojis to the app name, e.g., ``DALL·E'' into ``DALL·E\includegraphics[height=1em]{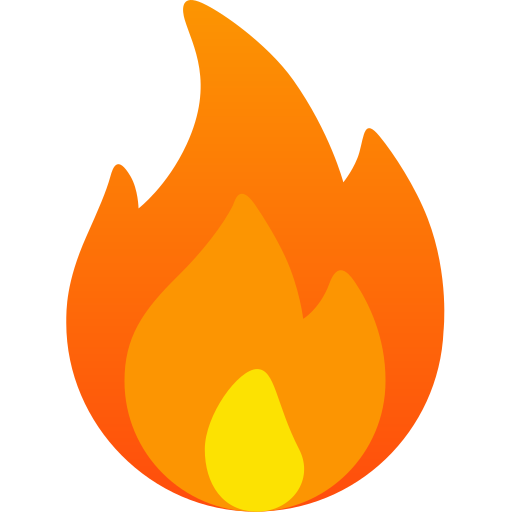}'' , exploiting the visual appeal and perceived legitimacy conveyed by emojis.

    \item \textit{String Rearrangement}: Rearranging parts of the package name, e.g., ``DALL·E'' to ``E·DALL''.
\end{enumerate}

\begin{figure}[h!]
    \centering
    \includegraphics[width=1\linewidth]{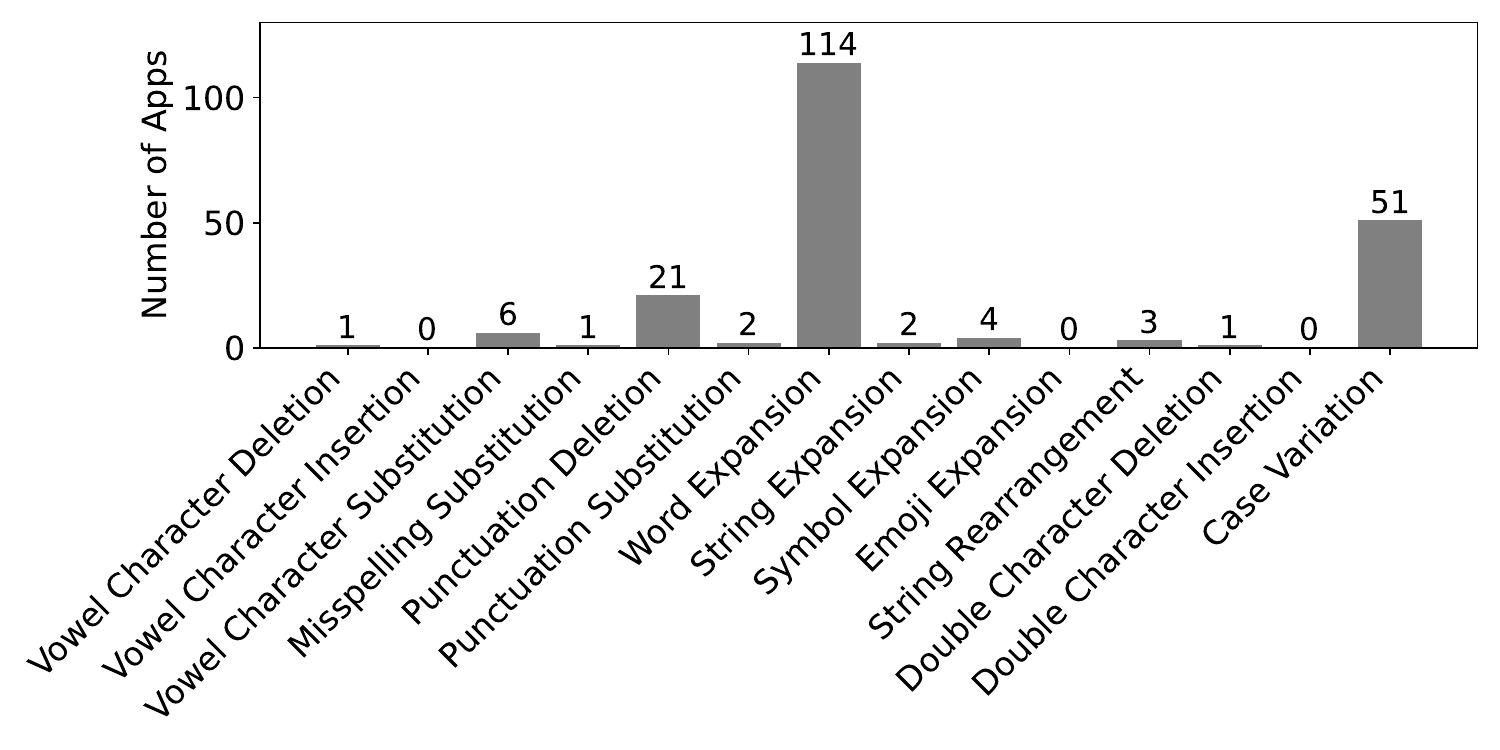}
    \caption{The distribution of squatting apps across models.}
    \label{fig:top10_virants_type}
\end{figure}

\noindent\textbf{Evaluation.} 
To evaluate the effectiveness of squatting generation models, we compare the results from our tool, \tool{}, with those of the traditional domain squatting approach, AppCrazy, used in the motivating study (see \autoref{sec:motivating}). The same set of 10 popular apps is used. With \tool{}, 14,520 squatting names are generated (as shown in Columns 4-5 of \autoref{tab:motivation}). Consistent with the process in the motivating study, these squatting names are searched in the GPTZoo dataset. The search identifies 206 squatting LLM app candidates with distinct IDs. 
\autoref{fig:top10_virants_type} shows the confirmed squatting apps generated by the 14 squatting-generation models. The newly added word expansion model is the most effective, with 114 squatting apps falling into this category. 

\subsection{Cloning Detection Models}
\label{subsec:cloning-detection}

We employed \textbf{Levenshtein distance} and \textbf{BERT-based semantic similarity} to detect plagiarism or app cloning in LLM app \texttt{descriptions} and \texttt{instructions}. Levenshtein distance identified exact or near-exact matches by measuring minimal edits, while the BERT model captured deeper semantic similarities, even with different wording. 
By analyzing both these components, we effectively detected cloning attempts, revealing instances of content replication ranging from direct copying to subtle paraphrasing, and highlighting the prevalence of cloning in the LLM app ecosystem.

\noindent\textbf{\textit{1) Levenshtein distance calculation}}

To detect cases of content cloning with minor variations, we employed Levenshtein distance algorithm~\cite{yu2007Levenshtein}, which \textbf{calculates the minimum number of single-character edits (insertions, deletions, or substitutions)} required to transform one string into another. For each app pair, we computed the Levenshtein distance between their \texttt{instructions} fields, which act as the core content or behavioral guide of the LLM app, similar to the source code. 

\begin{small}
\begin{equation}
\text{Levenshtein Similarity} = 1 - \frac{\text{Levenshtein Distance}}{\text{Maximum String Length}}
\end{equation}
\end{small}

\noindent where the \textbf{Maximum String Length} is the length of the longer string. This allowed us to compare app pairs with different text lengths. 
We focused on app pairs where the Levenshtein similarity scored between 0.95 and 1.0, excluding exact matches (similarity = 1). For example, with an \texttt{instructions} field of 500 characters, fewer than 25 modifications (5\% of the total length) would flag potential plagiarism, and for fields of 1000 characters, fewer than 50 changes would trigger detection. This threshold effectively captured minor variations while avoiding false positives due to insignificant changes.
To ensure the rigor of our analysis, we excluded comparisons where the \texttt{instructions} field was shorter than 50 characters, filtering out trivial entries such as single words or short phrases. This ensured that our analysis focused on substantial content replication. Focusing on high-similarity pairs enabled us to detect apps with minimal textual differences, suggesting potential attempts to clone content while avoiding exact duplication.

\noindent\textbf{\textit{2) BERT-based semantic similarity calculation}}

\begin{figure*}[h!]
    \centering
    \includegraphics[width=0.95\linewidth]{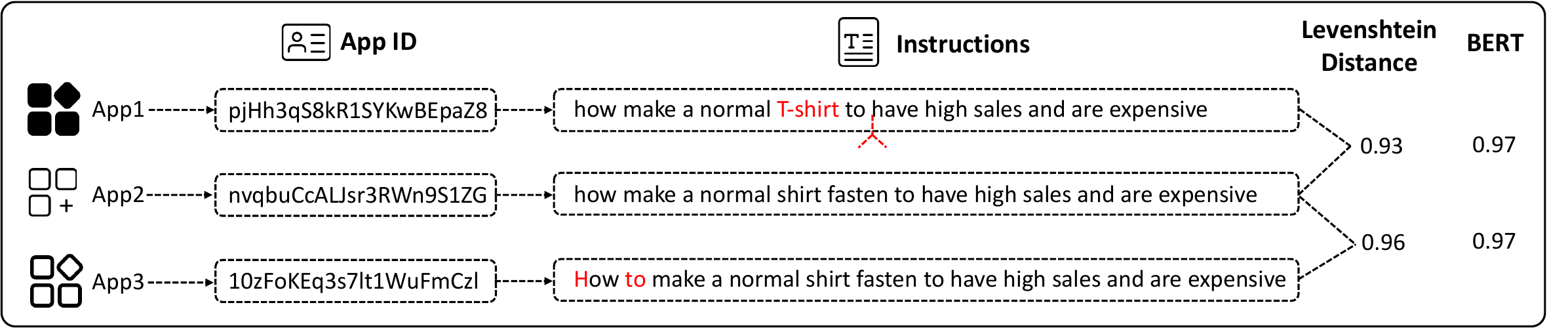}
    \caption{A real-world example highlights the differences between Levenshtein and BERT-based semantic similarity methods. Although all three apps convey the same core meaning, a typographical error with the term ``fasten'' in App2 and App3 causes the Levenshtein method to detect similarity only between these two, missing the similarity between App1 and App2.}
    \label{fig:real-example}
\end{figure*}

To detect more nuanced instances of app cloning, where the wording might vary while the underlying meaning remains consistent, we employed a BERT-based model~\cite{2021bert} to compute semantic similarity. Unlike character-based methods, this model \textbf{utilizes contextual embeddings to capture the semantic closeness between two pieces of text}, allowing for the detection of deeper, more subtle forms of copying.
The BERT model maps each input text into a high-dimensional vector space, where semantically similar texts have closer vector representations. Given two texts, \( t_1 \) and \( t_2 \), their semantic similarity score is calculated using the cosine similarity of their vector embeddings:

\begin{equation}
\text{Cosine Similarity}(t_1, t_2) = \frac{ \mathbf{v}_1 \cdot \mathbf{v}_2 }{ \| \mathbf{v}_1 \| \| \mathbf{v}_2 \|}
\end{equation}

\noindent where \( \mathbf{v}_1 \) and \( \mathbf{v}_2 \) are the embedding vectors generated by the BERT model for texts \( t_1 \) and \( t_2 \), respectively. The cosine similarity score ranges from 0 to 1, with higher values indicating greater semantic similarity.

We set a threshold of 0.95 for semantic similarity, meaning that if two texts scored above this value, they were flagged as having a strong semantic resemblance. This high threshold ensures precision, reducing the likelihood of false positives, while still capturing relevant instances of duplication.
Similar to the Levenshtein distance method, we excluded LLM apps where the \texttt{instructions} fields were shorter than 50 characters. Additionally, due to model limitations, we excluded LLM apps with \texttt{instructions} fields that exceeded 512 bytes in length. Unlike the Levenshtein method, however, we did not exclude app pairs with identical \texttt{instructions} fields, as these cases still provided valuable insights into semantic consistency.

\textbf{When the text's meaning remained consistent but the wording varied, the BERT-based approach was more effective than character-based methods.} For example, consider three apps, as shown in \autoref{fig:real-example}. 
The Levenshtein method misses the similarity between App1 and App2 due to minor text variations, while the BERT model effectively captures the semantic consistency across all three apps, demonstrating its advantage in detecting deeper similarities.

\section{Measuring Impersonation apps}
\label{result}

In this section, we use \tool{} to analyze impersonation apps in LLM app stores, focusing on squatting and cloning. Our investigation is guided by the following RQs:

\noindent\hangindent=2.5em\hangafter=1\textbf{RQ1 To what extent are squatting apps present? Do they primarily target popular apps?} 
We aim to analyze the prevalence of squatting apps in LLM app stores and determine whether they target more popular apps.

\noindent\hangindent=2.5em\hangafter=1\textbf{RQ2 How widespread is cloning apps, as another form of impersonation, in LLM app stores?} 
The low barrier to creating LLM apps has allowed cloning apps in LLM app stores to emerge. Our goal is to investigate the prevalence of these apps and understand their potential impact on users and the ecosystem.

\noindent\hangindent=2.5em\hangafter=1\textbf{RQ3 How many cases of potential cross-platform plagiarism exist? What are the situations in different stores?} 
This RQ aims to understand how app duplication across platforms impacts the uniqueness and integrity of LLM apps, and whether certain stores are more vulnerable to this issue than others.

\subsection{RQ1: Distribution of Squatting LLM Apps.}

In response to RQ1, we explore the prevalence and characteristics of app squatting among LLM apps. Our experiments rely on data from GPTs APP~\cite{GPTsappio}, the largest third-party GPT store, which provides rankings for the \textbf{top 1000 LLM apps}. This platform is essential for our analysis as it offers a ranking system not available in the official GPT Store~\cite{chatGPT}, making it a representative source.
To refine the results and minimize false positives, we applied a filtering process. Apps signed by the same developer but with slight name variations, such as platform-specific versions, were excluded. For example, different releases of an ``Image Generator'' app by the same developer across platforms were not considered squatting. Additionally, apps with common, non-branded names, like ``Image Generator'', were filtered out unless their package names followed predefined squatting patterns.

Once the data was extracted, we systematically compared it against the GPT dataset to identify instances of name duplication. This comparison revealed that 7,119 apps shared their names with those found in the top 1000 apps, suggesting a widespread occurrence of potential app squatting behavior. Notably, the most frequently duplicated app name was ``Prompt Engineer''~\cite{Prompt-Engineer}, which appeared 214 times across different records and was ranked 137th, indicating its significant popularity and the possible intent to capitalize on its recognition.  \autoref{tab:data-top5} below provides an overview of the five apps with the highest number of duplicate names, offering insights into the scale of this phenomenon and the types of apps most often targeted.

\begin{table}[!htbp]
    \centering
    \caption{Top 5 apps with the most duplicate names.}
    \resizebox{1\linewidth}{!}{\begin{tabular}{llc}
        \toprule[1.2pt]
        \textbf{App Name} & \textbf{Author Name} & \textbf{\# Duplicate Name Apps} \\
        \midrule
        Prompt Engineer & aitoolreport.com & 214 \\
        Translator & Caleb Ye & 154 \\
        Research Assistant & Liseli akayombokwa & 129 \\
        Resume Builder & masterinterview.ai & 127 \\
        Logo Creator & None & 116 \\
        \bottomrule[1.2pt]
    \end{tabular}}
    \label{tab:data-top5}
\end{table}

\begin{table}[!htbp]
    \centering
    \caption{Top 5 apps with the most squatting app names.}
    \resizebox{1\linewidth}{!}{\begin{tabular}{llc}
        \toprule[1.2pt]
        \textbf{App Name} & \textbf{Author Name} & \textbf{\# 
        Squatting Name Apps} \\
        \midrule
        AI Homework Helper & solvely.ai & 132 \\
        GPT Store Finder & EmbedAI & 126 \\
        Study+ Homework Helper & smartprompt.xyz & 122 \\
        Essay writing assistant & Corine Gorczany & 109 \\
        Python & Nicholas Barker & 106 \\
        \bottomrule[1.2pt]
    \end{tabular}}
    \label{tab:data-top5-squatting}
\end{table}

\begin{figure}
    \centering
    \includegraphics[width=1\linewidth]{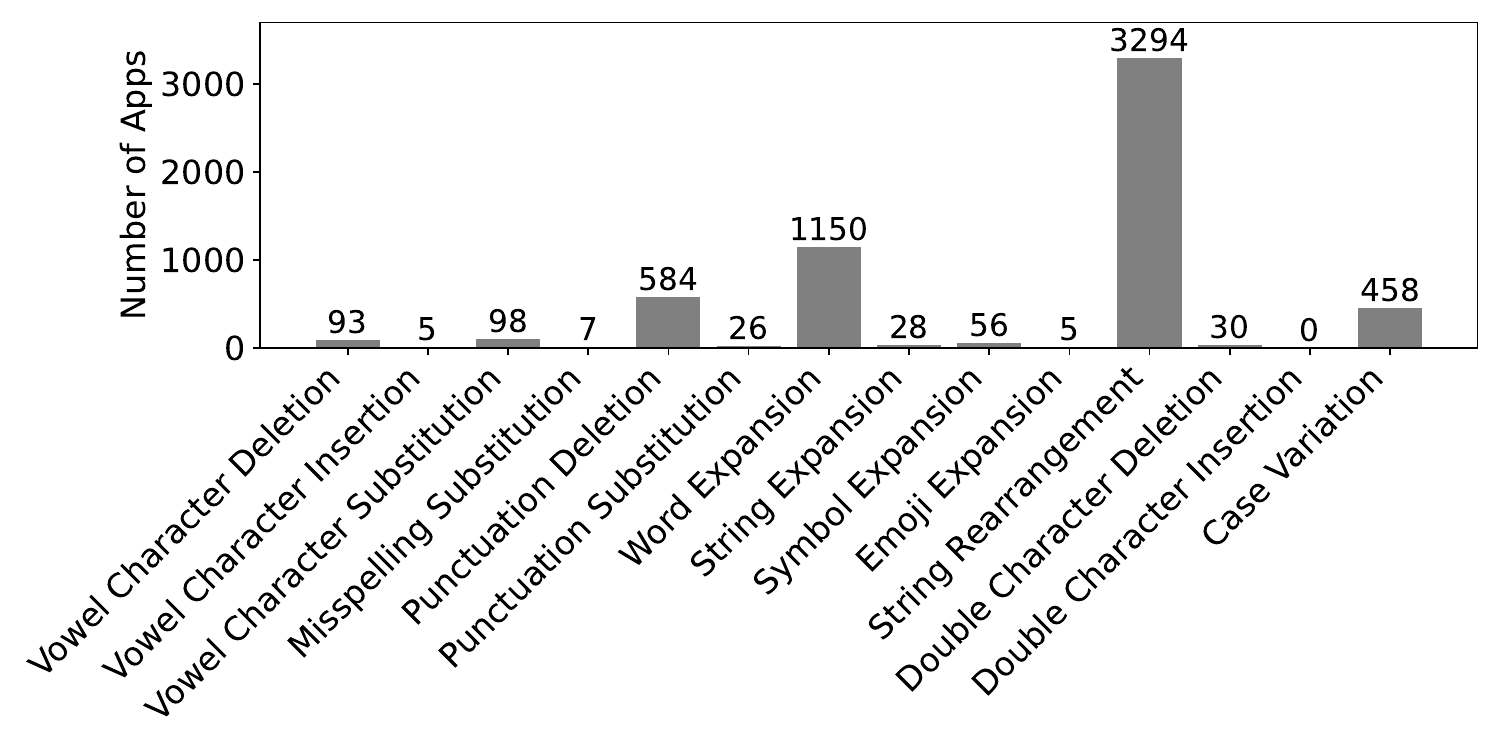}
    \caption{The distribution of squatting apps across models.}
    \label{fig:top1000_virants_type}
\end{figure}

To further examine the prevalence of app squatting, we utilized our tool \tool{}, to generate various name variations for the top 1000 apps, incorporating common squatting tactics such as case changes, character substitutions, misspellings, and expansions. Using these generated variants, we identified a total of 5,187 apps within the dataset that matched the modified names, highlighting the extensive use of squatting tactics. \autoref{tab:data-top5-squatting}
lists the top 5 apps with the most number of squatting apps. \autoref{fig:top1000_virants_type} shows the distribution of 5,834 squatting apps across 14 models. The top three patterns, \textbf{string rearrangement} (3,294), \textbf{word expansion} (1,150), and \textbf{punctuation deletion} (584), were newly introduced or modified for LLM apps, proving their effectiveness. Less common patterns like \textbf{case variation} (458 apps)  highlight additional attack strategies, offering insights for improving detection.

\begin{table*}[h!]
\centering
\caption{Overview of cloning apps in six LLM app stores.}
\resizebox{0.85\linewidth}{!}{
\begin{threeparttable}
\begin{tabular}{cccccccccccc}
\toprule[1.2pt]
\textbf{Store Name} & \textbf{LLM Apps} & \multicolumn{2}{c}{\textbf{Identical Instructions}} & \multicolumn{2}{c}{\textbf{Identical Descriptions}} & \multicolumn{2}{c}{\textbf{Identical Both\tnote{1}}} \\
\cmidrule(lr){3-4} \cmidrule(lr){5-6} \cmidrule(lr){7-8} 
 & \textbf{\# LLM Apps}  & \textbf{\# LLM Apps} & \textbf{\% of Total} & \textbf{\# LLM Apps} & \textbf{\% of Total} & \textbf{\# LLM Apps} & \textbf{\% of Total} \\
\midrule
GPT Store      & 662,294  & 36   & 0.01\%  & 7,570 & 1.14\%  & 0   & 0        \\
FlowGPT        & 34,271   & 784  & 2.29\%   & 944   & 2.75\%  & 121 & 0.35\%  \\
Poe            & 16,544   & 185  & 1.12\%   & 210   & 1.27\%  & 76  & 0.46\%  \\
Coze           & 51,912   & 33   & 0.06\%   & 0     & 0       & 0   & 0         \\
Cici           & 13,060   & 1    & 0.01\%   & 1     & 0.01\%  & 0   & 0         \\
Character.AI   & 7,048    & 20   & 0.28\%   & 40    & 0.57\%  & 12  & 0.17\%   \\
\midrule
\textbf{Total} & 785,129  & 1,058 & 0.13\% & 8,765 & 1.12\% & 209 & 0.03\%  \\
\bottomrule[1.2pt]
\end{tabular}
\begin{tablenotes}
    \footnotesize
    \item[1] Identical Both: Number of LLM apps with identical descriptions and instructions.
\end{tablenotes}
\end{threeparttable}}
\label{tab:data-summary}
\end{table*}

\begin{figure}[h]
    \centering
    \includegraphics[width=0.95\linewidth]{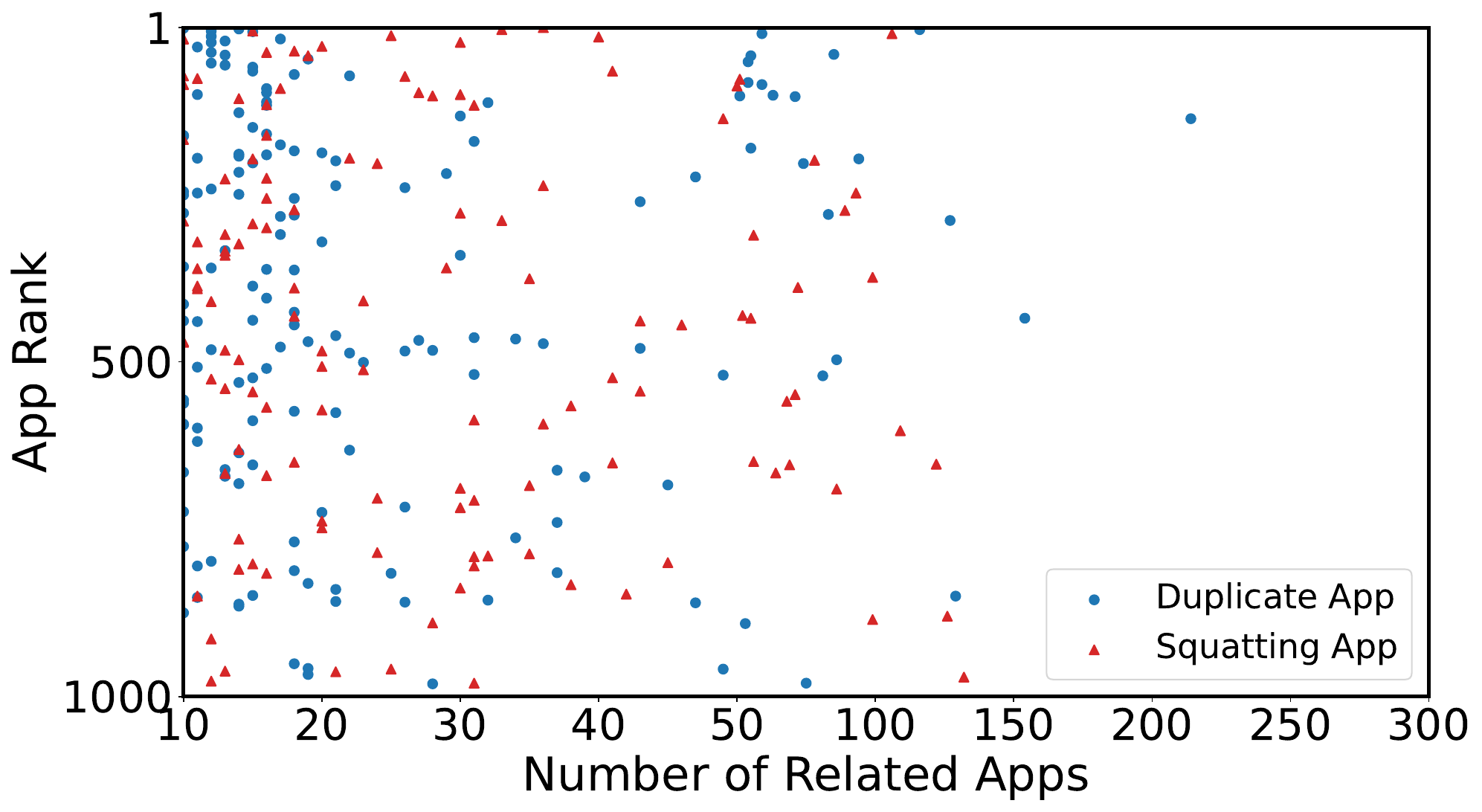}
    \caption{Distribution of fake and squatting apps by app rank.}
    \label{fig:top1000-squatting-distribution}
\end{figure}

To explore whether squatting apps specifically target more popular LLM apps, we analyzed the distribution of duplicate and squatting apps across different ranks. As shown in \autoref{fig:top1000-squatting-distribution}, higher-ranked apps (closer to the top of the y-axis) have more duplicate and squatting instances, indicated by the denser clustering in the upper region. 

\textbf{As LLM app stores target the general public, most app names are common and familiar, with few uniquely distinctive ones.} However, squatting still occurs among these app names. 
For instance, ``logogpts.cn'' created an app named ``LOGO'', and another app, ``LOGO+'', by ``Rodolfo Arce'', shares an identical description, suggesting potential squatting. 
This similarity strongly suggests a potential case of squatting. Nonetheless, we manually filtered out apps with very common names from the top 1000 apps. After filtering, we retained 654 apps and identified a total of 2871 squatting apps.

\vspace{0.2cm}
\noindent\underline{\textbf{Answer to RQ1.}}
We found that the top 1,000 LLM apps were associated with 5,834 squatting apps, with more popular apps being more frequently targeted. This could be due to their higher visibility and user demand. The most common method for generating squatting names in LLM app stores involves slight variations of the original app names, i.e. string Rearrangement, word Expansion,

\subsection{RQ2: Prevalence of Cloning LLM Apps.}

To address RQ2, we examined app cloning among 785,129 LLM apps from six platforms, focusing on two key fields: the \texttt{description} highlights the app's features, while the \texttt{instructions} serves as source code. We performed pairwise comparisons of these fields to identify identical or highly similar content, suggesting possible cloning. Our experiments covered both \textbf{exact matches} and \textbf{semantic similarities}, shedding light on the extent and nature of app cloning within the LLM app ecosystem.

\noindent\textbf{\textit{1) Exact match for identical content}}

We first used exact string matching to detect LLM apps with identical \texttt{instructions} or descriptions, effectively identifying direct duplicates where the text was copied verbatim, potentially misleading users into believing these apps are unique. Our analysis revealed significant app cloning across various LLM platforms, with 1,058 apps sharing identical \texttt{instructions} and 8,765 apps having identical descriptions. The GPT platform had the highest number of cloned descriptions (7,570 apps), while FlowGPT exhibited the most cloned \texttt{instructions} (784 apps). Additionally, 209 apps had both identical \texttt{instructions} and descriptions, with intra-platform plagiarism particularly common on platforms like FlowGPT and Poe. \autoref{tab:data-summary} (Columns 3-8) provides a detailed breakdown of these results across all platforms.

\noindent\textbf{\textit{2) Similarity detection}}

As detailed in \autoref{subsec:cloning-detection}, we used two methods: \textbf{Levenshtein distance} and\textbf{ BERT-based semantic similarity}, to detect app cloning where the \texttt{instructions} or descriptions were not identical but still highly similar. These approaches allowed us to identify subtle cloning behaviors, where minor textual changes were made to mask duplication.

\noindent\textbf{Levenshtein distance calculation.} Well-suited for detecting subtle variations like minor edits or typos, this method helps identify near-duplicate content. Applying a 0.95 similarity threshold to the \texttt{instructions} fields of 42,544 apps (after filtering out those with fewer than 50 characters), we identified 557 groups with high similarity, involving 1,637 apps. As shown in \autoref{tab:levenshtein}, FlowGPT had the most similar apps (1,396), with approximately 3.84\% of apps across the six platforms exhibiting near-duplicate \texttt{instructions}. These findings suggest widespread duplication and potential plagiarism, particularly on FlowGPT, warranting further investigation.

\begin{table}[!htbp]
    \centering
    \caption{Results of Levenshtein distance method.}
    \resizebox{1\linewidth}{!}{\begin{tabular}{lrrr}
        \toprule[1.2pt]
        \textbf{Store Name} & \textbf{Total Detections} & \textbf{Detection Results} & \textbf{Percentage} \\
        \midrule
        GPT Store           & 10,358 & 22 & 0.21\%  \\ 
        FlowGPT             & 23,906 & 1,396 & 5.84\% \\
        Poe                 & 5,177 & 188 & 3.63\% \\
        Coze                & 1,429 & 23 & 1.61\% \\
        Cici                & 0 & 0 & 0 \\
        Character.AI        & 1,674 & 13 & 0.78\% \\
        \midrule
        \textbf{Total}      & 42,544 & 1,637 & 3.84\% \\
        \bottomrule[1.2pt]
    \end{tabular}}
    \label{tab:levenshtein}
\end{table}

\noindent\textbf{BERT-based semantic similarity calculation.}  To further detect potential app cloning, we applied BERT-based semantic matching to the \texttt{instructions} fields, focusing on apps with 50 to 512 characters. This analysis covered 12,048 apps, using a similarity threshold of 0.95. We identified 253 groups of semantically similar apps, involving 2,113 apps. 
As shown in \autoref{tab:bert}, FlowGPT had the highest number of similar apps (1,705). BERT's ability to capture semantic meaning makes it effective for detecting cloning behaviors that go beyond exact text matches, revealing more nuanced forms of content duplication across platforms.

\begin{table}[!htbp]
    \centering
    \caption{Results of BERT-based method.}
    \resizebox{1\linewidth}{!}{\begin{tabular}{lrrr}
        \toprule[1.2pt]
        \textbf{Store Name} & \textbf{Total Detections} & \textbf{Detection Results} & \textbf{Percentage} \\
        \midrule
        GPT Store       & 1,930         & 92            & 4.77\%  \\ 
        FlowGPT         & 5,129         & 1,705         & 33.24\% \\
        Poe             & 3,092         & 258           & 8.34\% \\
        Coze            & 960           & 8             & 0.83\% \\
        Cici            & 0             & 0             & 0 \\
        Character.AI    & 937           & 50            & 5.34\% \\
        \midrule
        \textbf{Total}      & 12,048 & 2,113 & 17.54\% \\
        \bottomrule[1.2pt]
    \end{tabular}}
    \label{tab:bert}
\end{table}

\vspace{0.2cm}
\noindent\underline{\textbf{Answer to RQ2.}}
Our findings reveal a high prevalence of cloned apps across LLM app stores, with significant content duplication detected on multiple platforms. We identified 557 groups with highly similar \texttt{instructions} and 253 groups based on semantic similarity, involving thousands of apps. These clones pose risks by creating confusion over app authenticity, potentially undermining user trust and the integrity of the LLM app ecosystem.

\subsection{RQ3: Cross-platform Analysis}

\begin{figure}[h!]
    \centering
    \begin{subfigure}[b]{0.43\textwidth}
        \includegraphics[width=\textwidth]{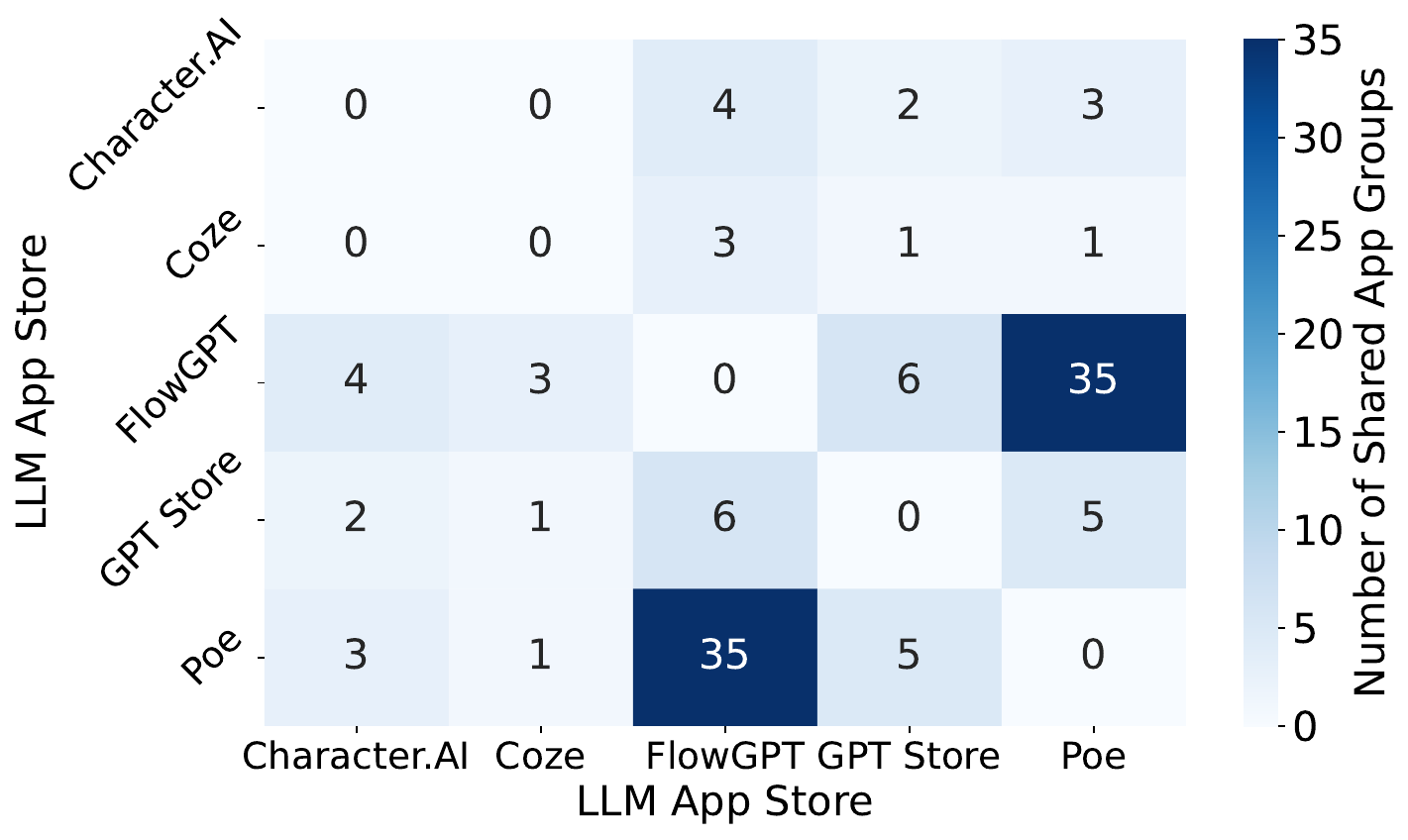}
        \caption{Cross-platform detection result of squatting.}
        \label{fig:cross_platform_2}
    \end{subfigure}
    \hfill
    \begin{subfigure}[b]{0.43\textwidth}
        \includegraphics[width=\textwidth]{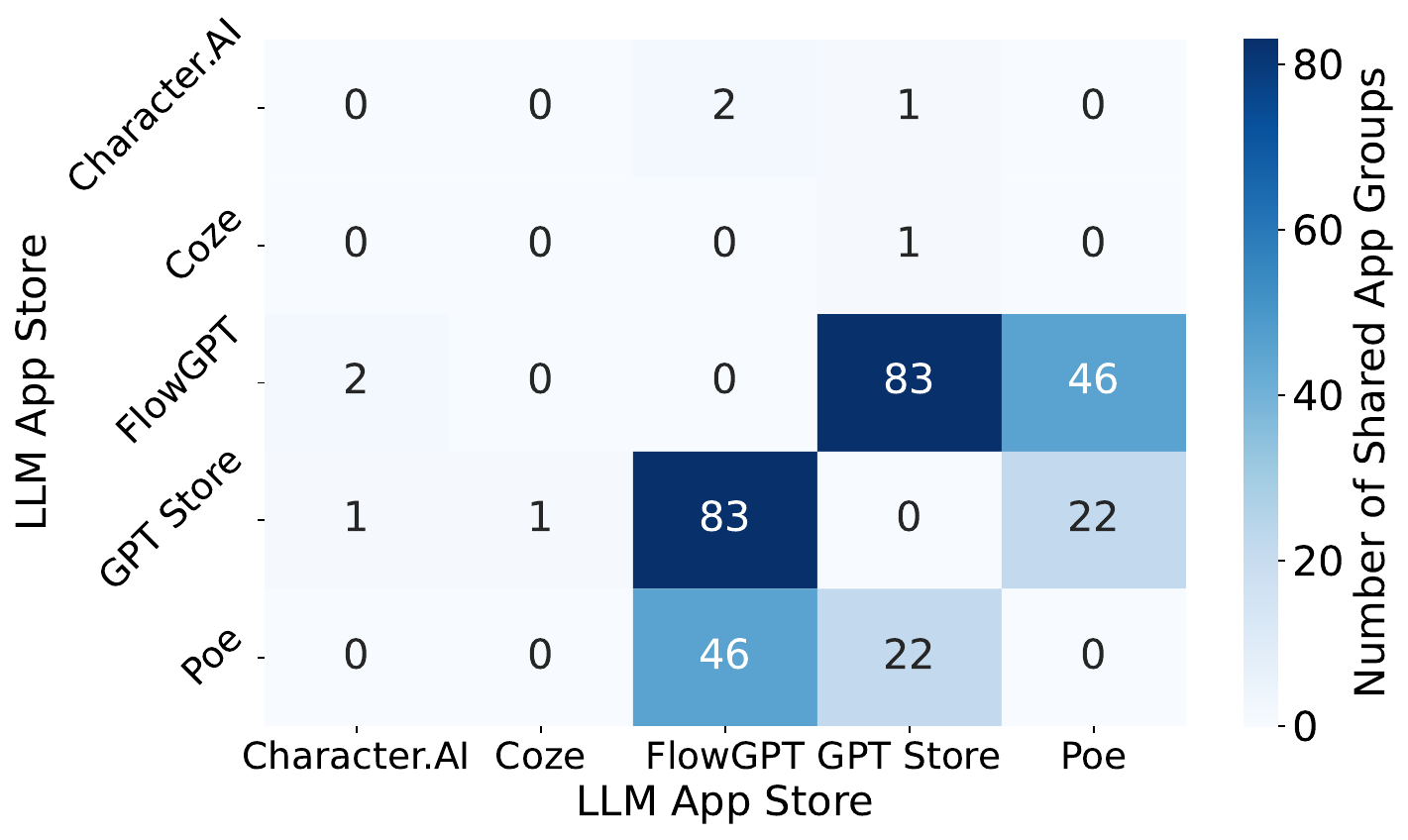}
        \caption{Cross-platform detection result of cloning.}
        \label{fig:cross_platform_1}
    \end{subfigure}
    \caption{Cross-platform detection result.}
    \label{fig:cross_platform}
\end{figure}

We analyzed app similarities across multiple LLM app stores to understand how duplication affects the uniqueness and integrity of LLM apps and whether certain stores are more vulnerable. By tracking platform information, we identified cross-platform plagiarism through app groups spanning different platforms. In the cloning experiment, we found 13 groups with identical \texttt{instructions}, 130 groups with identical descriptions, and 8 groups where both matched across platforms. Using the Levenshtein distance method, we identified 22 groups of suspected plagiarism, while BERT-based semantic matching revealed 40 groups with deep similarities, even when wording was altered. These findings highlight the complexity of cross-platform plagiarism, where cloning often involves subtle modifications preserving core content. The heatmaps in \autoref{fig:cross_platform_2} and \autoref{fig:cross_platform_1}  show that plagiarism is most common among FlowGPT, Poe, and GPT Store, with significant clustering suggesting these platforms are more prone to cloning and squatting compared to others.

\vspace{0.2cm}
\noindent\underline{\textbf{Answer to RQ3.}} Our analysis identified numerous cases of cross-platform plagiarism, with 13 groups sharing identical \texttt{instructions}, 130 groups with identical descriptions, and 8 groups matching in both. Additionally, 22 groups showed high similarity via Levenshtein distance, while BERT analysis found 40 groups with deep semantic overlap. FlowGPT, Poe, and GPT Store were particularly affected, suggesting these platforms are more prone to cloning and squatting, raising concerns about the integrity of LLM apps.

\section{Threat and Impact}
\label{result2}
we then examine the threat posed by impersonation apps and their impact on users and the LLM app ecosystem by exploring the following research questions:

\noindent\hangindent=2.5em\hangafter=1\textbf{RQ4 How many impersonation (squatting and cloning) apps are malicious?} 
Understanding how many of these squatting and cloning apps are malicious will provide insight into the extent of harm they can cause, such as spreading malware or conducting phishing attacks.

\noindent\hangindent=2.5em\hangafter=1\textbf{RQ5 What is the impact of these impersonation apps on users and the LLM app ecosystem?}
This RQ seeks to assess how impersonation apps affect user trust and security, as well as their broader impact on the LLM app ecosystem’s integrity.

\subsection{RQ4: Malware Presence}

\begin{figure}[h]
    \centering
    \includegraphics[width=0.95\linewidth]{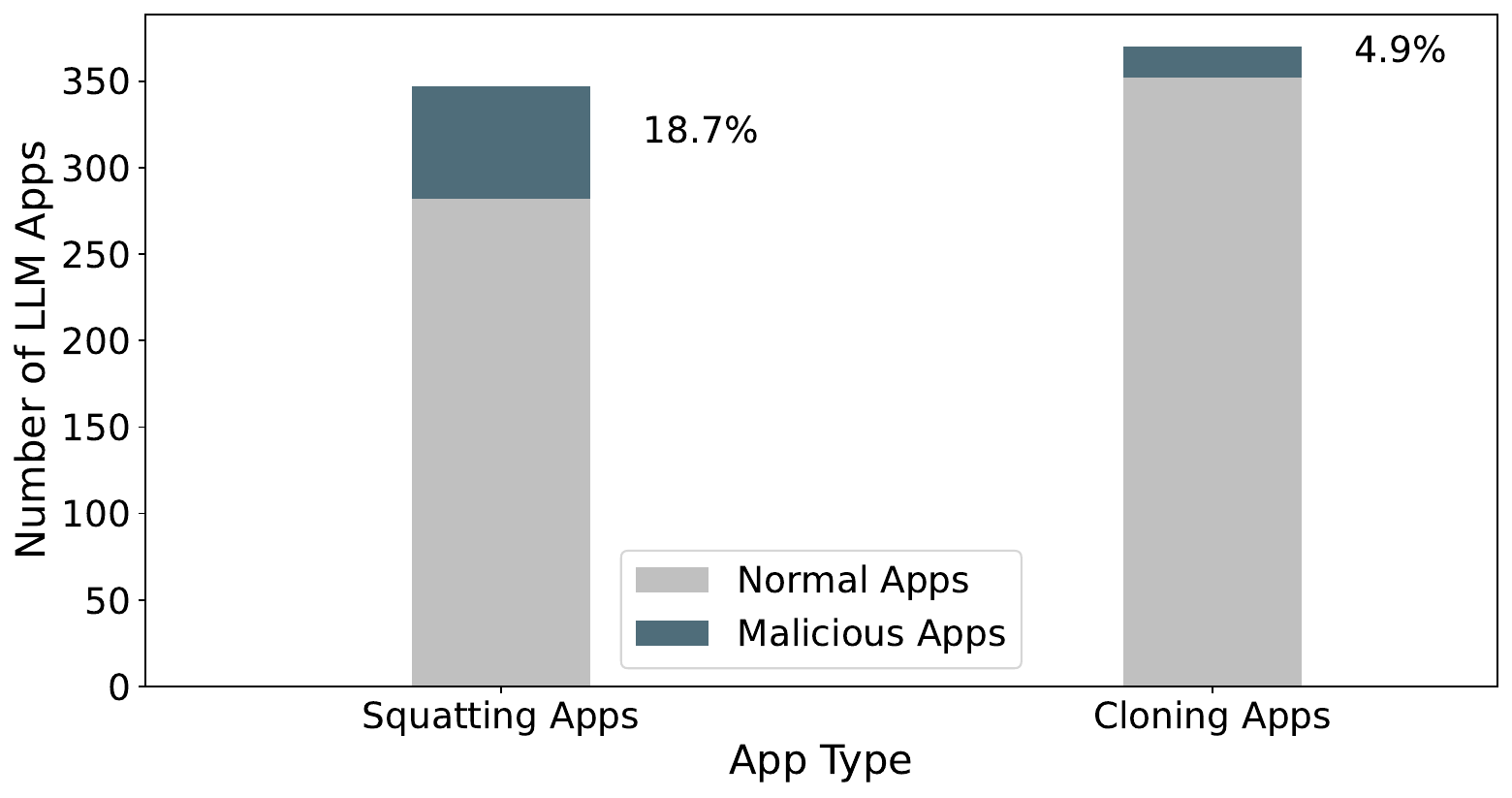}
    \caption{Result of sampling analysis.}
    \label{fig:RQ4_1}
\end{figure}

\textbf{When certain apps exhibit a very high degree of similarity in the fields of \texttt{app name}, \texttt{description}, and \texttt{instructions}, it is clear that these apps are deliberately imitating others}, strongly suggesting an intent to impersonate. To quantify this, we conducted a comprehensive analysis of the squatting and cloning experiment results from RQ1 and RQ2 and identified 227 apps that met the criteria for high similarity.
Following this, we aimed to evaluate the potential malicious behavior within squatting and cloning apps. Out of the 3,509 squatting apps and 9,575 cloning apps identified, we selected a representative sample of 347 squatting and 370 cloning apps, using a 95\% confidence level and a 5\% confidence interval to ensure statistical significance. This sample underwent manual inspection to detect malware, phishing, and ad injection, assessing the risks these impersonation apps might pose to users. ~\autoref{fig:RQ4_1} illustrates the proportion of malicious apps identified in the sample.

\begin{figure}[h]
    \centering
    \includegraphics[width=0.95\linewidth]{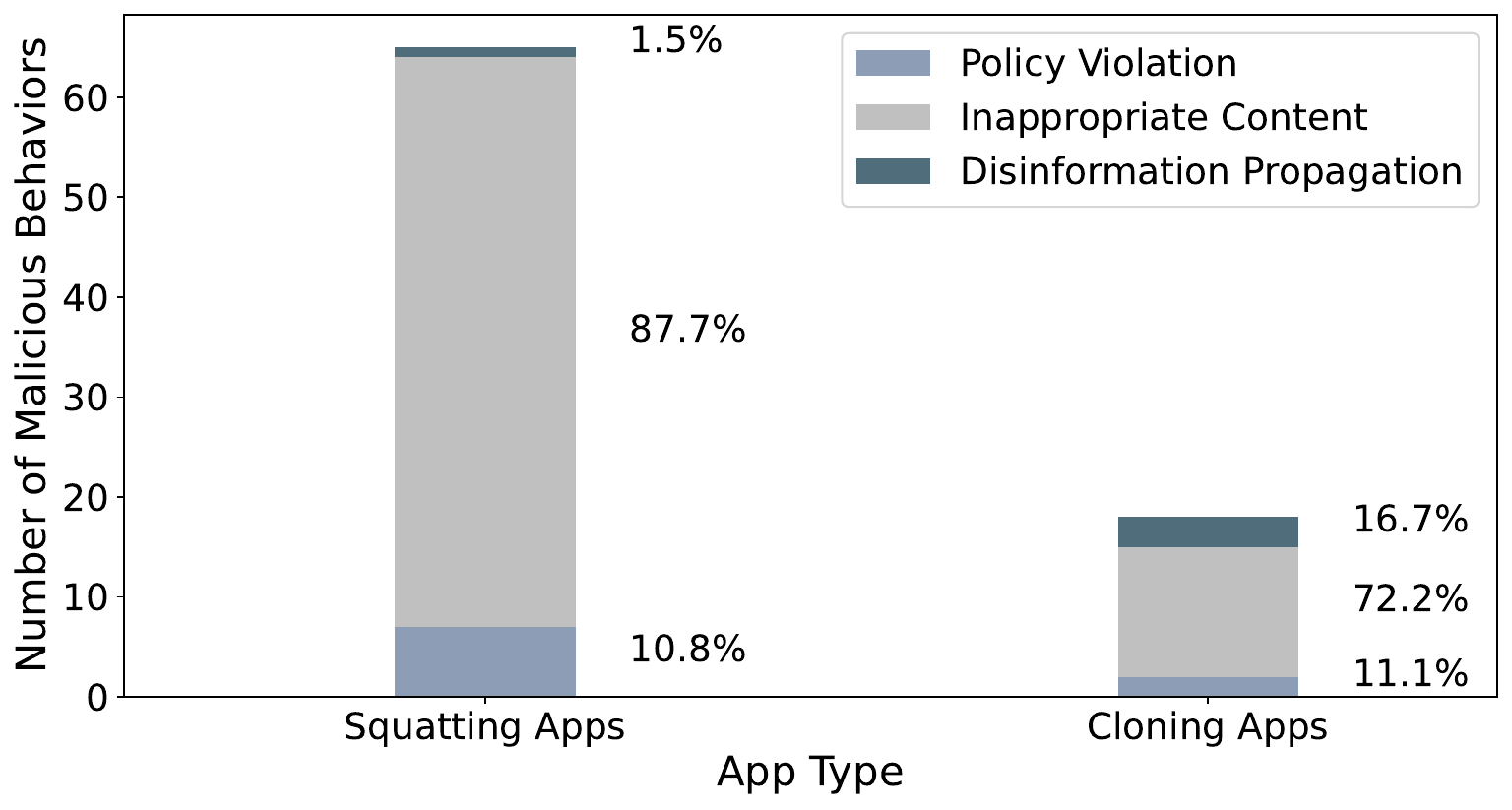}
    \caption{The distribution of malicious behaviors.}
    \label{fig:RQ4_2}
\end{figure}

After a thorough manual inspection of 347 selected squatting apps, we found 18.7\% violating LLM app usage policies~\cite{hou2024security}. Of these, 2\% provided \texttt{instructions} encouraging guideline violations, and 0.3\% linked to an unknown website, raising phishing concerns. Alarmingly, 16.4\% apps directed users to generate inappropriate content, including sexual, violent, or illegal material. In the 370 cloned apps, 4.9\% were non-compliant with LLM policies. Among them, 0.5\% encouraged violations, and 3.5\% promoted inappropriate content. Notably, 0.8\% exhibited fraudulent behavior, claiming to operate ``fully automated with a high win rate'' to lure users with false promises. As shown in \autoref{fig:RQ4_2}, the malicious behaviors detected in our study fall into three categories: \textbf{policy violations}, \textbf{inappropriate content}, and \textbf{disinformation}, with inappropriate content being the most prevalent. Apps promoting illegal content, misleading users, or encouraging policy violations pose serious risks to user safety and data security, undermining trust in LLM platforms and the app ecosystem. If left unchecked, these apps could normalize unethical practices and attract more malicious actors. Our findings highlight the urgent need for stricter regulations and robust monitoring in LLM app stores to ensure user protection and maintain ethical standards, fostering a secure and trustworthy environment.

\vspace{0.2cm}
\noindent\underline{\textbf{Answer to RQ4.}} 
We found that 227 apps exhibited high similarity in \texttt{app name}, \texttt{description}, and \texttt{instructions}, indicating deliberate impersonation. Additionally, among the examined apps, 65 out of 347 squatting apps and 18 out of 370 cloning apps were found to be non-compliant. These apps often provided \texttt{instructions} that violated policies, generated inappropriate content, or engaged in fraudulent practices, underscoring significant security risks and the urgent need for stronger regulations to protect users and the ecosystem.

\subsection{RQ5: Impact on Users}

\begin{figure}[h!]
    \centering
    \includegraphics[width=0.9\linewidth]{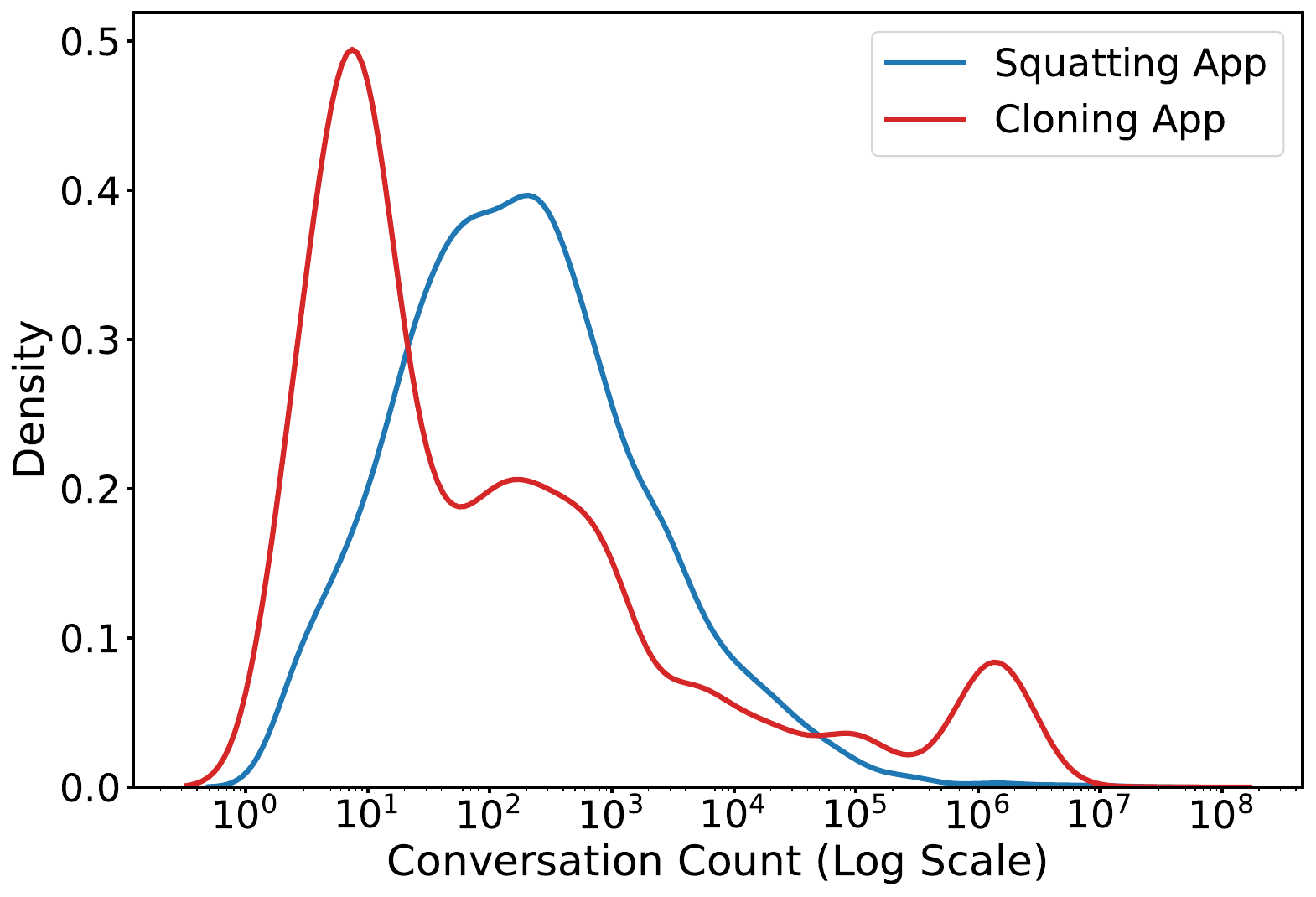}
    \caption{Kernel density distribution of conversation counts.}
    \label{fig:kde}
\end{figure}

Squatting apps in LLM app stores have reached high usage levels, significantly affecting users. Of the 3,509 identified squatting apps, 2,835 had conversation counts between 0 and 1,000, showing a large portion with lower engagement. However, 674 apps exceeded 1,000 conversations, and 50 surpassed 50,000, demonstrating substantial user interaction. In particular, the top squatting app had 12,969,368 conversations, while another app with nearly identical instructions ranked third with 4,236,464 conversations. These two apps, published by different creators, suggest potential unauthorized replication, posing risks due to high engagement.
For cloning apps, of the 9,575 identified, 7,828 had conversation counts between 0 and 1,000, while 1,747 recorded over 1,000, and 726 exceeded 100,000, highlighting significant user interaction. The top cloned app reached 27,527,998 conversations. \autoref{fig:kde} shows the distribution of conversation counts, with squatting apps peaking broadly at higher counts (around $10^2$ to $10^5$) and cloning apps peaking sharply at lower counts (around $10^1$), indicating that squatting apps generally achieve higher user engagement and visibility, thus posing a greater threat.

High conversation volumes in squatting and cloning apps increase the risk of users unknowingly interacting with unauthorized or low-quality replicas, exposing them to unverified content, potential malicious activities, and privacy issues. These counterfeit apps often lack updates and support, leading to a poorer user experience and overshadowing legitimate apps, complicating access to authentic resources and weakening trust in LLM app stores.

\vspace{0.2cm}
\noindent\underline{\textbf{Answer to RQ5.}} 
Squatting and cloning apps in LLM stores demonstrate high user engagement, significantly affecting user experience and platform integrity. Of 3,509 squatting apps, 674 had over 1,000 conversations, with the top app reaching 12.9 million interactions. Similarly, 1,747 of 9,575 cloning apps exceeded 1,000 conversations, with the most-used app hitting 27.5 million. Interaction with unauthorized replicas exposes users to security risks and diminishes visibility for legitimate apps, highlighting the need for stronger oversight.

\section{Discussion}
\label{discussion}

\subsection{Mitigation \& Implications}

We propose strategies to address the challenges of LLM app squatting and cloning,  focusing on three key stakeholders:

\noindent\textbf{LLM app store managers}. Platforms should enhance their app review processes by incorporating automated and manual checks to detect duplicate or similar apps. Advanced plagiarism detection tools can help identify potential plagiarism during the submission process. Additionally, recommendation algorithms should be improved to prioritize unique, high-quality content and reduce the visibility of cloned apps, ensuring that users encounter a wider variety of original options. 

\noindent\textbf{LLM app developers}. Developers should take an active role in protecting their apps from squatting and cloning. This includes selecting distinct, non-conflicting app names and regularly monitoring for potential infringements. If unauthorized replicas are found, developers should report these to the platform maintainers to ensure prompt action. 

\noindent\textbf{End users}. Educating users about the risks of cloned or unauthorized apps is crucial. They should be taught to identify suspicious apps and use tools to verify legitimacy. Developers and platforms can help by offering resources like tutorials and reports to guide users in avoiding squatting attacks and choosing legitimate apps.

\subsection{Threat to Validity}
\label{subsec:limitation}

\noindent\textbf{Identical app name detection.} Unlike traditional mobile app squatting detection, our approach includes identical app names in LLM app stores, where duplicates are allowed. Squatting attackers tend to use exact names to mimic legitimate apps and deceive users. Including identical names helps detect as many squatting apps as possible. As many  developers choose names casually, this can lead to unintentional duplication and false positives. To better distinguish intentional squatting from accidental duplication, we combine squatting and cloning detection based on both name and instruction similarity.

\noindent\textbf{Popular app selection.} Our detection of LLM app squatting focuses mainly on the GPT Store, as it is the only platform with app ranking data. This research targets popular apps, which we believe is appropriate since attackers tend to focus on well-known applications. However, future work will examine how to generalize our findings to more LLM apps.

\noindent\textbf{Tool limitation.} Although \tool{} is specifically tailored for LLM apps, the generation model may still be incomplete, leaving room for other complex squatting methods. To address this, we designed the squatting generation models in \tool{} as an easily extensible tool, allowing new patterns to be added seamlessly. In cloning detection models, due to input length limitations, we only analyzed instructions of a specified length, potentially missing cloning in apps with longer instructions. However, our results still provide initial evidence of cloning in the LLM app ecosystem, and we plan to improve our detection methods in the future.

\noindent\textbf{Cross-platform deduplication} Different authors may use different names across platforms, and in our cross-platform plagiarism analysis, we can only accurately identify cases where the author names are identical. This limitation highlights the need for additional verification processes to distinguish between legitimate cross-platform distribution and unauthorized replication by third parties.

\section{conclusion}
\label{conclusion}

In this study, we conducted the first large-scale analysis of LLM app squatting and cloning using our tool, \tool{}. Through the detection of 14 squatting generation techniques and leveraging both Levenshtein distance and BERT-based semantic analysis, we identified over 5,000 squatting apps from variations of top app names. Across six major platforms, we found 3,509 squatting apps and 9,575 cloning cases. Our sampling revealed that 18.7\% of the squatting apps and 4.9\% of the cloning apps exhibited malicious behavior, highlighting significant risks to user security and the integrity of LLM app stores. These findings underscore the need for stronger oversight and protective measures in the LLM app ecosystem.

\section*{Data availability}
The artifact is publicly accessible at \url{https://anonymous.4open.science/r/LLM_app_squatting_and_cloning-6000/}.

\bibliographystyle{IEEEtranS}
\bibliography{main}

\end{document}